\documentclass[runningheads]{llncs}

\usepackage{eccv}

\usepackage{eccvabbrv}

\usepackage{graphicx}
\usepackage{booktabs}
\usepackage{multirow}
\usepackage{geometry}
\usepackage[accsupp]{axessibility}  

\definecolor{cvprblue}{rgb}{0.21,0.49,0.74}
\definecolor{yujieColor}{rgb}{0,0.4,0}

\newcommand{\methodName}{CanoVerse}

\usepackage{hyperref}

\usepackage{orcidlink}

\usepackage{makecell}
\usepackage{tabularx} 
\usepackage{booktabs} 

\begin{document}

\title{CanoVerse: 3D Object Scalable Canonicalization and Dataset for Generation and Pose} 

\titlerunning{CanoVerse}

\author{Li Jin$^{1*}$ \and
Yuchen Yang$^{1*}$ \and
Weikai Chen$^{2\dagger}$ \and
Yujie Wang$^{3\dagger}$ \and
Dehao Hao$^{4}$ \and
Tanghui Jia$^{5}$ \and
Yingda Yin$^{2}$ \and
Zeyu HU$^{2}$ \and
Runze Zhang$^{2}$ \and
Keyang Luo$^{2}$ \and
Li Yuan$^{5}$ \and
Long Quan$^{4}$ \and
Xin Wang$^{2}$ \and
Xueying Qin$^{1\dagger}$
}

\authorrunning{L. Jin, Y. Yang, W. Chen et al.}

\institute{$^{1}$SDU, $^{2}$LIGHTSPEED, $^{3}$UNC Chapel Hill, $^{4}$HKUST, $^{5}$PKU}


\maketitle

\footnotetext{$\text{*}$Equal contribution.}
\footnotetext{$\dagger$Corresponding author.}

\begin{abstract}
  3D learning systems implicitly assume that objects occupy a coherent reference frame. Nonetheless, in practice, every asset arrives with an arbitrary global rotation, and models are left to resolve directional ambiguity on their own. This persistent misalignment suppresses pose-consistent generation, and blocks the emergence of stable directional semantics. To address this issue, we construct \methodName{}, a massive canonical 3D dataset of 320K objects over 1,156 categories -- an order-of-magnitude increase over prior work. At this scale, directional semantics become statistically learnable: \methodName{} improves 3D generation stability, enables precise cross-modal 3D shape retrieval, and unlocks zero-shot point-cloud orientation estimation even for out-of-distribution data. This is achieved by a new canonicalization framework that reduces alignment from minutes to seconds per object via compact hypothesis generation and lightweight human discrimination, transforming canonicalization from manual curation into a high-throughput data generation pipeline. The \methodName{} dataset will be publicly released upon acceptance. Project page: \url{https://github.com/123321456-gif/Canoverse}
\end{abstract}

\section{Introduction}


A persistent source of uncontrolled variation in 3D vision is \textit{orientation}. Web assets, 3D scans, and 3D generative models all appear in arbitrary unknown frames. While scale and translation are routinely normalized, orientation is often ignored, yet it systematically degrades representation stability. 
This is not a minor nuisance: orientation variation could fragment shape retrieval (the same instance becomes dozens of rotated identities), confuse generative models (global pose becomes inconsistent and uncontrollable), and contaminate cross-category evaluation (generalization cannot be measured fairly).
Crucially, directional semantics like ``front'' or ``up'' cannot be discovered from geometry alone. They only become learnable priors when they appear consistently across \textit{many} canonicalized instances. In other words, canonical orientation must exist at \textit{dataset scale} before 3D models can internalize reliable orientation-consistent 3D semantics.

\begin{figure}[t]
    \centering
    \includegraphics[width=.89\linewidth]{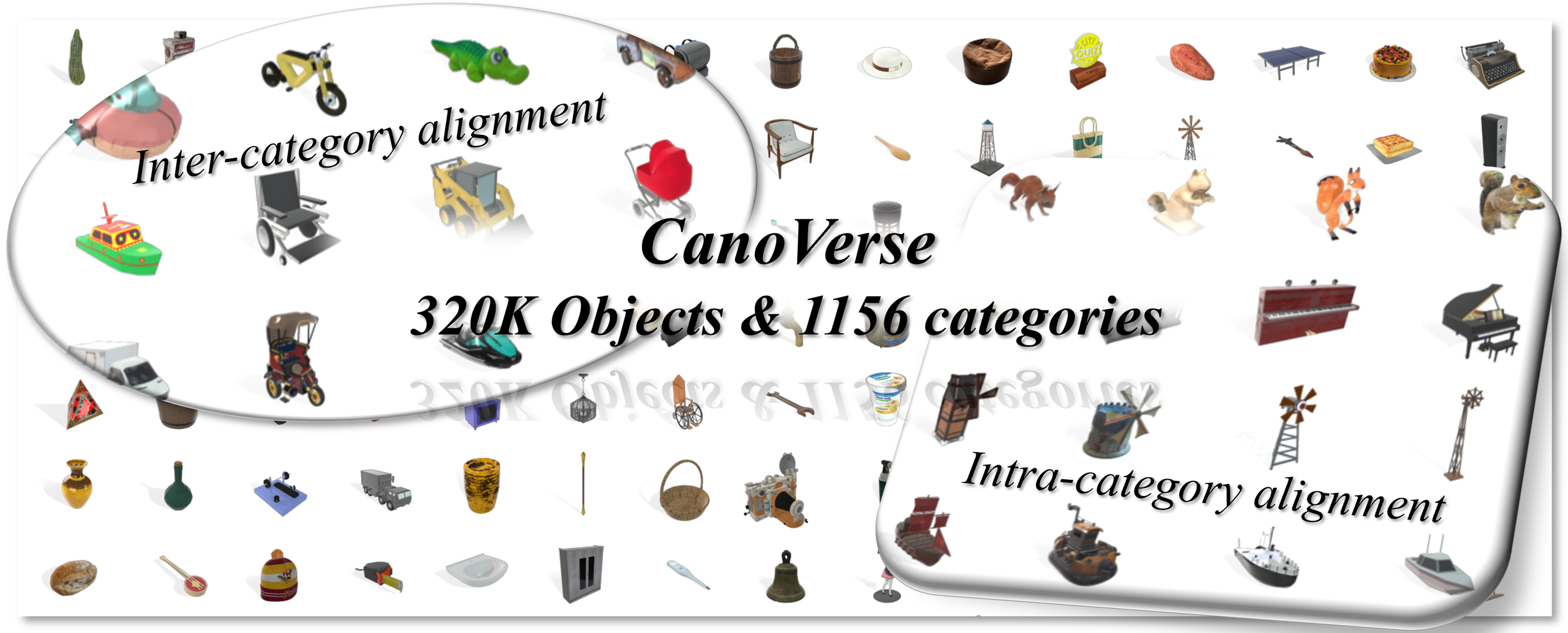}
    \vspace{-2mm}
    \caption{We present CanoVerse, a large-scale canonical dataset standardizing object orientation, size, and position with intra- and inter-category alignment. It contains 1,156 categories and 320k objects, making it the largest canonical dataset to date, an order of magnitude larger than existing ones.}
    \label{fig:teaser}
    \vspace{-1mm}
\end{figure}


However, existing canonical 3D datasets remain limited in scale: 32K (COD~\cite{jinli2025}), 14.8K (Objaverse-OA~\cite{lu2025orientationmatters}),  and 6K (OmniObject3D~\cite{wu2023omniobject3d}) -- numbers that are two orders of magnitude smaller than the raw 3D assert universe. The reason is simple: current canonicalization workflows still rely heavily on manual alignment, making it prohibitively expensive to scale up the dataset. Consequently, canonical orientation today exists only in small pockets of data, and is far too sparse to support robust generalization studies or to serve as a foundation for orientation-consistent 3D representation learning.

To address these limitations, we introduce a scalable canonicalization framework that dramatically reduces human cost of alignment: manual annotation time per object drops from minutes to seconds. 
In particular, we design a tailored pipeline that reformulates alignment as hypothesis generation plus lightweight human discrimination. 
Unlike prior efforts that primarily rely on geometry heuristics or fully manual alignment, Our framework jointly leverages \textit{geometry} and \textit{semantic} cues, and designs category-specific judgment criteria for both horizontal and vertical dimensions, to generate a small set of canonical pose candidates that include the ground-truth canonical pose. 
This hybrid strategy preserves the accuracy of human judgment while eliminating the search over $\mathrm{SO}(3)$, yielding substantially higher throughput with high coverage and accuracy. 

\begin{figure}[t]
    \centering
    \includegraphics[width=.89\linewidth]{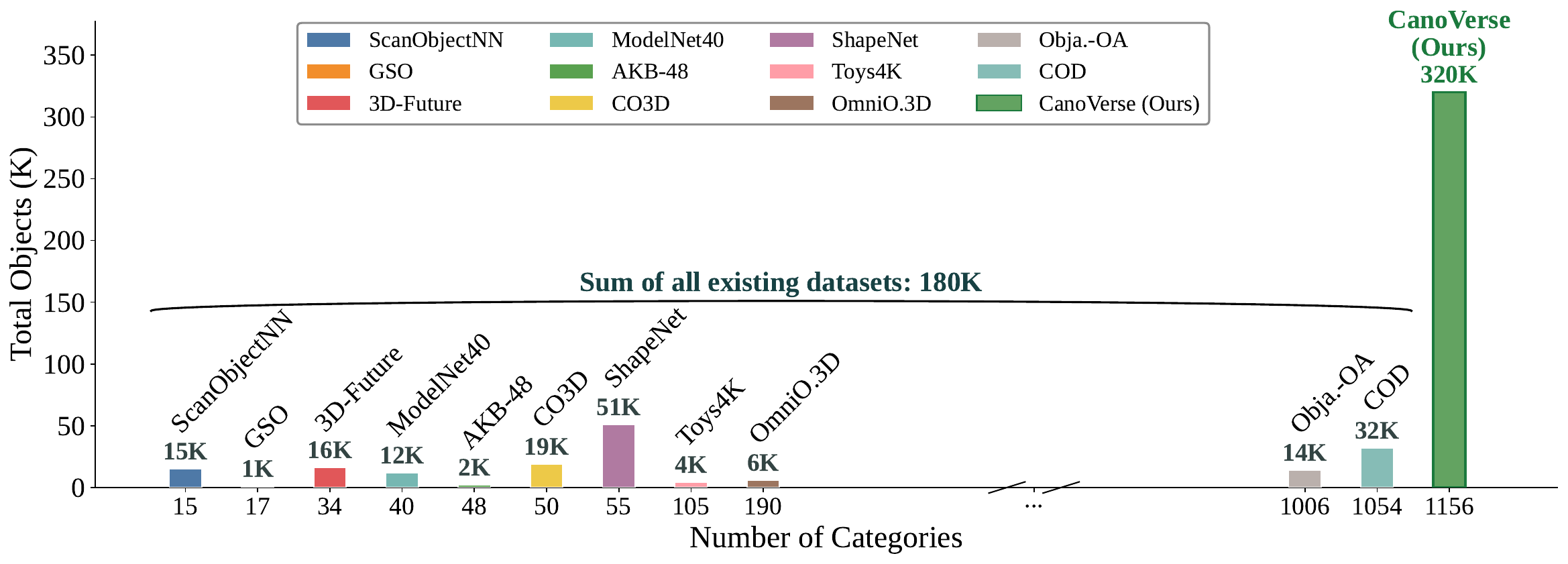}
    \vspace{-3mm}
    \caption{Scale and category coverage of different canonical datasets. Our dataset boasts the largest number of categories and objects among existing canonical datasets—far surpassing each individual one and even the total of all aforementioned datasets. }
    \label{fig:dataset_compare}
    \vspace{-1mm}
\end{figure}

With this engine in place, we canonicalize the Objaverse and Objaverse-XL collection and construct \textit{\methodName{}} -- \textbf{320K canonicalized objects across 1,156 categories} -- which, to our knowledge, is the largest canonical 3D dataset to date (See Figure \ref{fig:dataset_compare}). 
\methodName{} provides sufficient statistical mass to make directional semantics learnable across broad coverage. 
At this scale, directional semantics become statistically learnable. 
We observe substantial gains across multiple tasks, including improved point-cloud orientation estimation with stronger generalization to unseen objects and real-world scanned out-of-distribution data, more orientation-stable 3D generation when models are trained on canonical data, and more accurate cross-modal 3D shape retrieval. 
Finally, \methodName{} exhibits higher canonicalization quality than COD~\cite{jinli2025} and Objaverse-OA~\cite{lu2025orientationmatters} (the largest and latest canonical dataset with a comparable number of covered categories) under quantitative comparison, and a user study confirms that our pipeline reduces annotation time from minutes to seconds per object while maintaining alignment reliability. 
We summarize our contributions as follows:
\begin{itemize}
    \item A scalable canonicalization framework. We propose a novel pipeline that fuses geometric and semantic cues, designs multiple canonicalization criteria, generates canonical candidates, and converts 3D object canonical annotation from 3D continuous optimization to 2D selection, cutting per-object manual annotation time from minutes to seconds.

    \item A large-scale canonical 3D dataset coded \methodName{} with 320K objects across 1,156 categories -- an order-of-magnitude increase over prior canonical datasets. We will release \methodName{} upon publication.
    \item Substantial gain on downstream tasks. \methodName{} improves 3D generation stability, scales to cross-modal 3D shape retrieval, and, critically, enables zero-shot point cloud orientation estimation, which was previously impractical due to insufficient canonical data.
\end{itemize}

\section{Related Work}

\paragraph{\textbf{3D Canonicalization Datasets.}}
Early 3D canonical datasets such as ShapeNet~\cite{shapenet2015}, ModelNet40~\cite{Zhirong15CVPR}, and OmniObject3D~\cite{wu2023omniobject3d} have made significant contributions to related research. However, these datasets are limited in scale: for example, ShapeNet contains approximately 51k objects across 55 categories, ModelNet40 includes 12k models spanning 40 categories, and OmniObject3D comprises 6k objects covering 190 categories. Such scales remain insufficient to meet the data demands of large-model training, making it difficult to enable directional semantics to be learnable across broad category coverage.
The key reason for the difficulty in large-scale expansion of these datasets lies in their heavy reliance on manual alignment, which incurs high alignment costs. The COD~\cite{jinli2025} dataset adopts a one-shot method for object alignment, eliminating manual costs but suffering from poor alignment quality.
In contrast, the method proposed in this paper can ensure high alignment quality while significantly reducing manual alignment costs: it shortens the alignment time from the minute level to the second level, greatly improving data annotation efficiency and thus enabling the expansion of larger-scale datasets.

\paragraph{\textbf{Object Orientation Estimation.}}
A feasible approach for object orientation estimation is to leverage 6D object pose estimation methods, which can be broadly divided into two categories. 
Instance-level pose estimation~\cite{wen2024foundationpose,peng2019pvnet,sun2022onepose,lee2025any6d} predicts relative poses based on known 3D CAD models or reference images. 
However, in practical scenarios, such models or references are often unavailable, limiting the applicability of these methods.
To improve generalization, category-level pose estimation methods~\cite{wang2019normalized,chen2020category,lin2022category,fan2022object} learn shape priors shared within a category. Additionally, Orient Anything\cite{orient_anything} trains an object orientation estimation model using large model-processed data.
While more flexible, they require large-scale orientation-aligned 3D datasets, which are costly to construct and thus remain scarce.
In contrast, our method enables the creation of large-scale canonicalized datasets covering diverse categories and instances, allowing category-level pose estimation methods to scale beyond the limited category coverage of existing approaches.


\paragraph{\textbf{3D Object Generation.}}
Recent advances in 3D generation have largely followed the latent VAE-diffusion paradigm~\cite{dora,craftsman,hunyuan2,wu2024direct3d}, inspired by the success of 2D diffusion models~\cite{rombach2022high}.
Current frameworks can be broadly grouped into VecSet-based approaches~\cite{3dshape2vecset,clay,craftsman,hunyuan2}, which encode shapes as latent vector sets and decode them via neural implicit functions, and sparse-voxel-based approaches~\cite{xcubes,trellis,wu2025direct3d}, which use structured voxel grids for explicit spatial reasoning.
While these systems achieve impressive geometric and visual fidelity, they remain sensitive to inconsistencies in data orientation and geometry.
As training assets often lack a consistent canonical frame, generative models tend to produce unstable poses and duplicated symmetric parts, highlighting the need for large-scale, orientation-consistent data supervision.
By providing 320K canonicalized objects with consistent global orientation, \methodName{} can directly address this limitation, enabling 3D generators to internalize directional priors and produce pose-stable and coherent results.

\section{Method}

\begin{figure}[t]
    \centering
    \includegraphics[width=.97\textwidth]{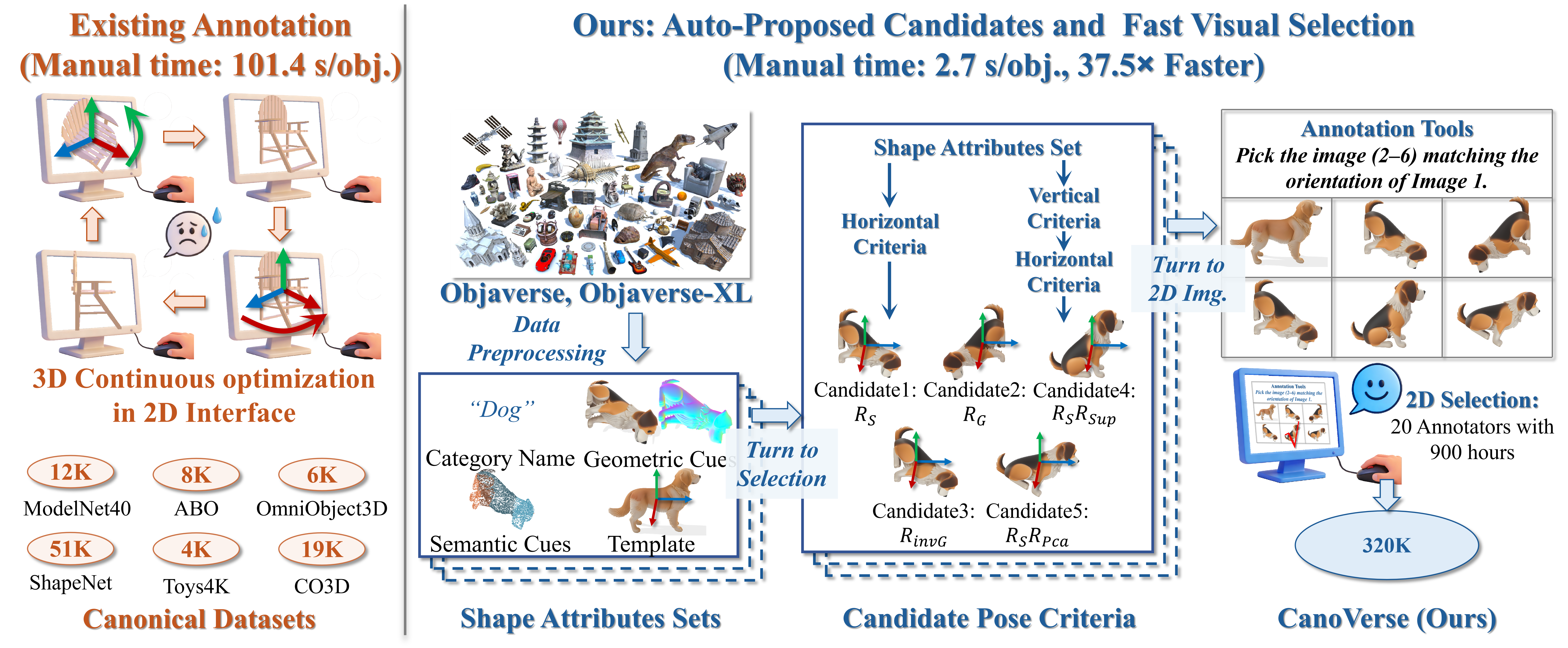}
    \vspace{-3mm}
    \caption{3D canonicalization pipeline. Left: prior manual alignment ($\sim$101.4\,s/object) yields small-scale datasets. Right: our candidate generation ($\sim$3\,s) plus one-click selection ($\sim$2.7\,s) gives $\sim$37.5$\times$ faster annotation and enables a 320K canonical dataset.}
    \label{fig:method_overview}
    \vspace{-1mm}
\end{figure}

We develop a two-stage pipeline for large-scale, category-level canonicalization (Figure~\ref{fig:method_overview}): 
\textbf{(I) candidate-pose generation} (\Cref{sec:stageI}) and \textbf{(II) interactive selection} (\Cref{sec:stageII}).
Stage~I operates in a \emph{one-shot, category-level} manner: given a single template per category, it automatically produces a compact set of reliable pose candidates per instance.
Stage~II provides an annotator-friendly GUI for one-click selection among the top-$5$ candidates produced from Stage~I. This design amortizes effort while preserving accuracy at scale.

\subsection{Candidate-Pose Generation}
\label{sec:stageI}

\

\noindent \textbf{Data Preprocessing.}
We extract geometric and semantic cues, per-category templates, and (for symmetric objects) symmetry axes and angles. For each shape, we sample a colored point cloud and normalize its translation and scale to a unit sphere, then use Uni3D~\cite{zhou2023uni3d} for category assignment and Find3d~\cite{ma20243d} for semantic part segmentation; the per-category semantic text used in segmentation follows COD~\cite{jinli2025} and is obtained via large language models.
%

For each category, we choose a representative template as the canonical frame, with orientation following human convention—for a bicycle, the z-axis is front and the y-axis is up. We harmonize across categories so that related ones share aligned axes; the front of cars and bicycles, for instance, both map to the z-axis. For symmetric objects, following NOCS~\cite{wang2019normalized}, we define a symmetry axis and rotational symmetry angle per category (Figure~\ref{fig:sym_anno}, left), shared by all instances; defining these once per category keeps annotation tractable. Extended preprocessing details are in the supplementary material.


\subsubsection{Candidate Pose Criteria.}
\begin{figure}[t]
    \centering
    \includegraphics[width=.85\linewidth]{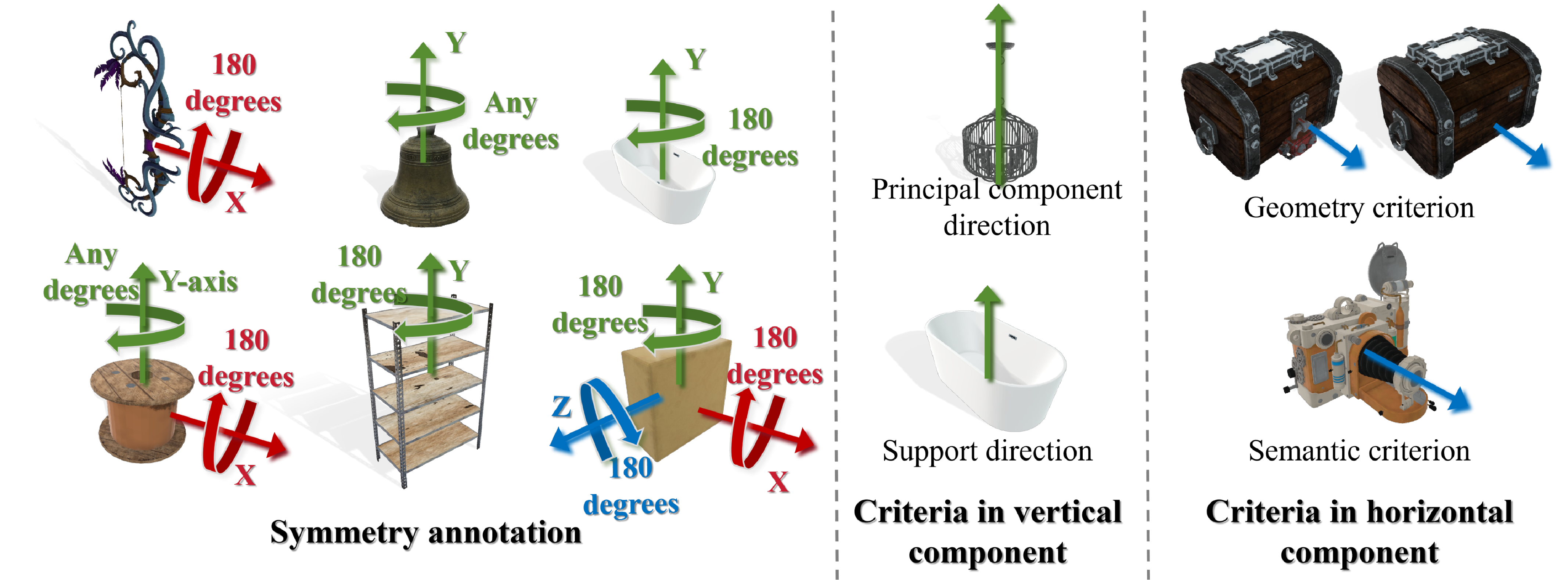}
    \vspace{-3mm}
    \caption{For symmetric categories, we annotate the symmetry axis and angle (left). Due to diverse object shapes, unified annotation is impractical. We thus design separate standards based on object characteristics and generate candidate poses in the vertical (middle) and horizontal (right) directions.}
    \label{fig:sym_anno}
    \vspace{-1mm}
\end{figure}

Humans align objects based on different cues depending on the category. For example, cameras are oriented according to semantic features such as lens direction, while cartons are aligned based on geometric structure. Consequently, there is no universal rule for canonicalizing 3D object orientations across diverse categories.
To ensure that the canonical pose is covered within a candidate set, we design multiple alignment criteria capturing both geometric and semantic cues commonly used in human perception.
We decouple object orientation into horizontal and vertical components (Fig.~\ref{fig:sym_anno}, middle and right) and define separate alignment criteria for each. Candidate poses are then generated by applying these criteria  independently or sequentially to the input object. These criteria form the basis of our candidate pose generation stage.

\paragraph{\textbf{Criteria in Vertical Component.}}
Vertical canonicalization with respect to gravity generally falls into two categories. The first aligns objects using a supporting surface, while the second relies on principal component analysis (PCA). For example, a cup can be canonicalized by placing it stably on a horizontal plane, whereas swords or chandeliers require alignment via suspension.

For the first category, we compute candidate supporting surfaces that allow the object to remain in static equilibrium. For a rigid body, the center of mass must lie within the supporting polygon~\cite{Upright08}. We therefore enumerate faces satisfying this condition and restrict to poses where the supporting face is parallel to the ground. The resulting candidate poses are rendered, and a vision-language model (VLM) selects the one that best matches human expectations for upright orientation. The corresponding rotation is denoted as $\boldsymbol{R}_\mathrm{Sup}$. Implementation details are provided in the supplementary material.

For the second category, we align the object to the category template using Principal Component Analysis (PCA). 
Specifically, we compute the first two principal axes of both the object and the template from the covariance eigendecomposition of their centered point clouds. 
As PCA eigenvectors are sign-ambiguous, the object axes admit four polarity combinations relative to the template axes, yielding four candidates:
\begin{equation}
\small
\boldsymbol{R}_i =
\begin{bmatrix}
\boldsymbol{v}_1^t & \boldsymbol{v}_2^t & \boldsymbol{v}_1^t \!\times\! \boldsymbol{v}_2^t
\end{bmatrix}
\begin{bmatrix}
\sigma_{i1}\boldsymbol{v}_1^o &
\sigma_{i2}\boldsymbol{v}_2^o &
\sigma_{i1}\boldsymbol{v}_1^o \!\times\! \sigma_{i2}\boldsymbol{v}_2^o
\end{bmatrix}^{\!\top},\, i \in \{1,2,3,4\},
\end{equation}
where $\boldsymbol{v}_1^o,\boldsymbol{v}_2^o$ and $\boldsymbol{v}_1^t,\boldsymbol{v}_2^t$ denote the first two principal axes of the object and template point clouds, respectively, and $\sigma_{i1},\sigma_{i2}\in\{+1,-1\}$ resolve the sign ambiguity of PCA eigenvectors.

We resolve this ambiguity by utilizing semantic cues to determine the correct polarity of each axis. The semantic part distributions of the object and template provide additional constraints for calibrating the object’s axial directions. The optimal rotation is selected by minimizing the semantic alignment error:
\begin{equation}
\boldsymbol{R}_\mathrm{Pca} = \arg\min_{\boldsymbol{R}_i}\ \frac{1}{m}\sum_{k=1}^m\text{CD}\big(\mathcal{S}_k^t,\boldsymbol{R}_i\mathcal{S}_k^o\big).
\end{equation}
$m$ is the semantic part amount, $\mathcal{S}_k^o$ and $\mathcal{S}_k^t$ are the $k$-th semantic point sets of the object and template, and $\text{CD}(\cdot,\cdot)$ is the Chamfer distance. This selects the polarity combination that best aligns the semantic part layout with the template.

\paragraph{\textbf{Criteria in Horizontal Component.}} 
Similarly, we introduce two criteria for horizontal alignment, leveraging geometric and semantic cues to align object orientations with the category template.

Specifically, the geometry criterion mainly aligns the categories with a similar shape and topology. For instance, Benches share a strip-like geometric form in shape and a support structure composed of legs and a backrest frame in topology. Thereby, we evaluate the geometric similarity by calculating the chamfer distance between the object and the template along the horizontal direction, and obtain the alignment results by selecting the most similar candidate. The geometry criterion is:
$
\theta^* = \arg\min_{\theta\in[0,2\pi)}\ \text{CD}\big(\mathcal{P}^t,R_z(\theta)\mathcal{P}^o\big),
$
where $\theta$ is the horizontal rotation angle around the vertical axis $z$; $R_z(\theta)$ is rotation matrix for horizontal rotation; $\mathcal{P}^o,\mathcal{P}^t$ is point sets of the target object and template, respectively; $\text{CD}(\cdot,\cdot)$ is Chamfer Distance, which measures the geometric similarity between two point sets; $\theta^*$ is optimal horizontal rotation angle that minimizes the Chamfer Distance.
Substituting $\theta^*$ into $R_z(\theta)$, we obtain the geometric alignment matrix $\boldsymbol{R_G}$ and its 180° inverted version $\boldsymbol{R_{\mathrm{inv}G}}$ for symmetric objects:
\begin{equation}
\boldsymbol{R_G} = R_z(\theta^*), \quad \boldsymbol{R_{\mathrm{inv}G}} = R_z(\theta^*+\pi)
\end{equation}

\begin{figure}[t]
    \centering
    \includegraphics[width=.89\linewidth]{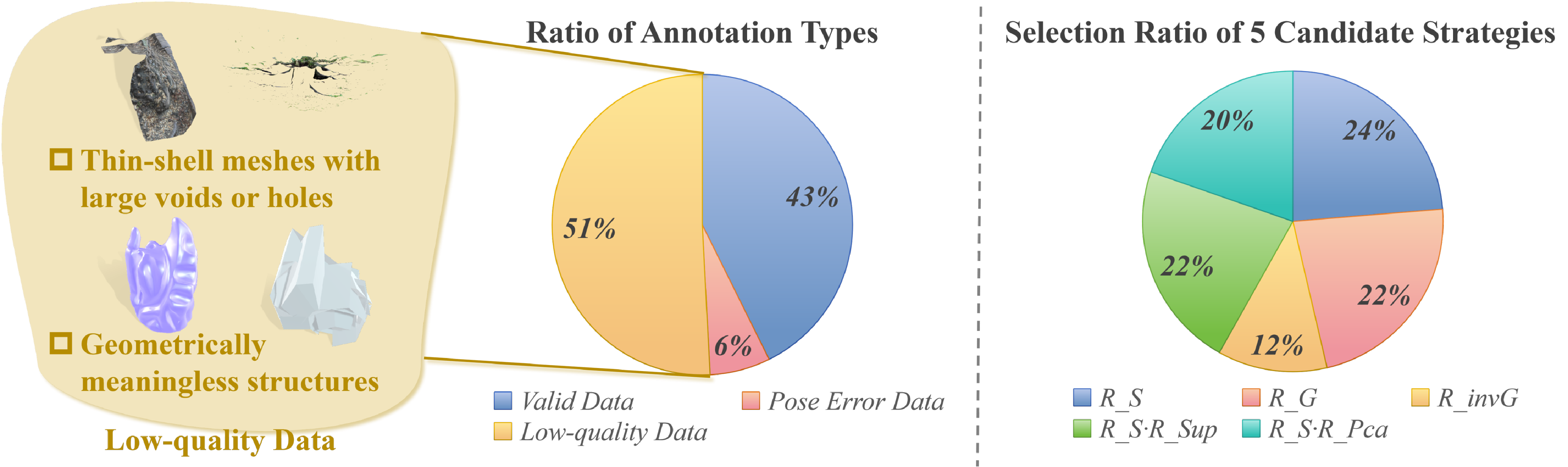}
    \vspace{-3mm}
    \caption{Annotation statistics. 750K annotated; 51$\%$ discarded for quality, 6$\%$ for pose errors, 43$\%$ retained (middle). Main discard types including thin-shell meshes, meaningless structures (left). Selection distribution over the 5 candidates (right).}
    \label{fig:annoCounts}
    \vspace{-5mm}
\end{figure}

In addition, in categories with significant shape diversity, the semantics criterion play a dominant role in guiding alignment. For instance, cameras can be aligned by lens’ orientation, a core semantic cue, regardless of whether they are instant cameras with miniaturized lenses or telephoto cameras with a bulky lens. Therefore, inspired by COD~\cite{jinli2025}, we use the joint energy function combined semantic and geometry to fully leverage the dominant semantic guidance while compensating for the lack of fine-grained geometric details in pure semantic alignment. The semantic criterion is defined as follows:
\begin{equation}
\begin{split}
\small
E_g(\theta) &= \text{CD}\big(\mathcal{P}^t,R_z(\theta)\mathcal{P}^o\big), \quad
E_s(\theta) = \frac{1}{m}\sum_{k=1}^m\text{CD}\big(\mathcal{S}_k^t,R_z(\theta)\mathcal{S}_k^o\big) \\
\theta^* &= \arg\min_{\theta\in[0,2\pi)} -\left( e^{-E_s(\theta)} \cdot \sum_{\omega_k \in \boldsymbol{\Omega}} \mathcal{N} \left( \theta \mid \omega_k, \mathbf{I} \right) \right)
\end{split}
\end{equation}
where $E_g$ and $E_s$ measure geometric and semantic fit (Chamfer distance on full point sets and on $m$ semantic part pairs under $R_z(\theta)$), and the objective favors angles near the geometric extrema of $E_g$. More details are in the supplementary material. We set $\boldsymbol{R_S} = R_z(\theta^*)$.

\subsubsection{Pose Generation Pipeline.}
Our canonical dataset is built from Objaverse and Objaverse-XL curated from public platforms such as Sketchfab. The orientation distribution in the raw data is non-uniform and reflects common human placement—animals are mostly upright with minor facing ambiguity, while objects like swords appear either horizontal or vertical. Based on these traits and the vertical/horizontal criteria above, we form five candidate poses per object so that the human-preferred canonical pose lies among them.
The five candidates are: (i)~\emph{horizontal semantic}: $\boldsymbol{R_S}$; (ii)~\emph{horizontal geometric}: $\boldsymbol{R_G}$; (iii)~\emph{horizontal geometric, 180° flip} for symmetric categories such as beds: $\boldsymbol{R_{\mathrm{inv}G}}$; (iv)~\emph{vertical (support surface) then horizontal semantic}: $\boldsymbol{R_S}\boldsymbol{R}_{\mathrm{Sup}}$; (v)~\emph{vertical (PCA) then horizontal semantic}: $\boldsymbol{R_S}\boldsymbol{R}_{\mathrm{Pca}}$. Together they cover horizontal-only and vertical-then-horizontal alignments.

\begin{figure}[t]
    \centering
    \includegraphics[width=.89\linewidth]{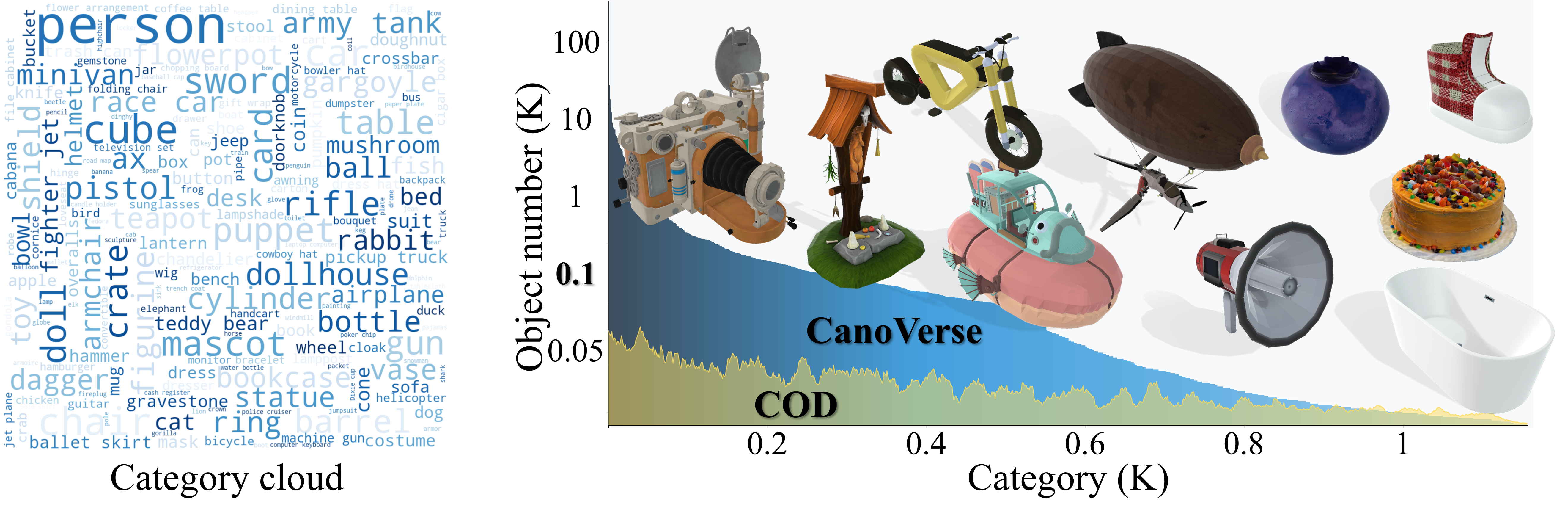}
    \vspace{-4mm}
    \caption{
    Category coverage (left) and long-tailed distribution with more shapes per category than COD (right). Vertical axis: linear below 0.1, exponential above.}
    \label{fig:long-term}
\end{figure}

\subsection{Interactive Selection}
\label{sec:stageII}
\textbf{Task Formulation.} During annotation, annotators perform a discrete visual selection task. For each instance, annotators are presented with the reference canonical template and renderings under 5 candidate rotations, and are asked to select the best-aligned one.

\noindent \textbf{Annotation Protocol.} To ensure consistent understanding, each annotator completes a short calibration before large-scale annotation; we monitor per-category consistency and conduct lightweight rechecking for low-confidence categories. Annotators discard instances that are misclassified or exhibit severe geometric degradation such as substantial mesh loss or incomplete structure. Protocol details, including calibration setup and workforce scale, are given in the supplementary material.

\noindent \textbf{Quality Control and Annotation Statistics.} Across 750K annotated samples, 320K are retained after screening; the average time per instance is under 5 seconds. The selection distribution over the five candidates (Fig.~\ref{fig:annoCounts}) shows that all candidate types contribute to final alignment decisions. Our final dataset covers a large number of categories with a long-tailed distribution (Fig.~\ref{fig:long-term}) and enriches most categories compared to COD~\cite{jinli2025}. 

\section{Experiments}


We evaluate \methodName{} on three downstream tasks: 3D object orientation estimation (\Cref{sec:pose_est}), 3D object generation (\Cref{sec:3dgen}), and 3D shape retrieval (\Cref{sec:shape_ret}). Two additional studies assess the efficiency and quality of our annotation pipeline (\Cref{sec:anno_val}).

\subsection{3D object Orientation estimation} \label{sec:pose_est}
In 3D object orientation estimation, the objective is to recover the rotation that aligns an arbitrarily orientation object to the canonical space. We evaluate orientation estimation under three settings to validate dataset quality, cross-dataset transferability, and performance scaling with increasing data volume.

\begin{figure}[t]
    \centering
    \includegraphics[width=.89\linewidth]{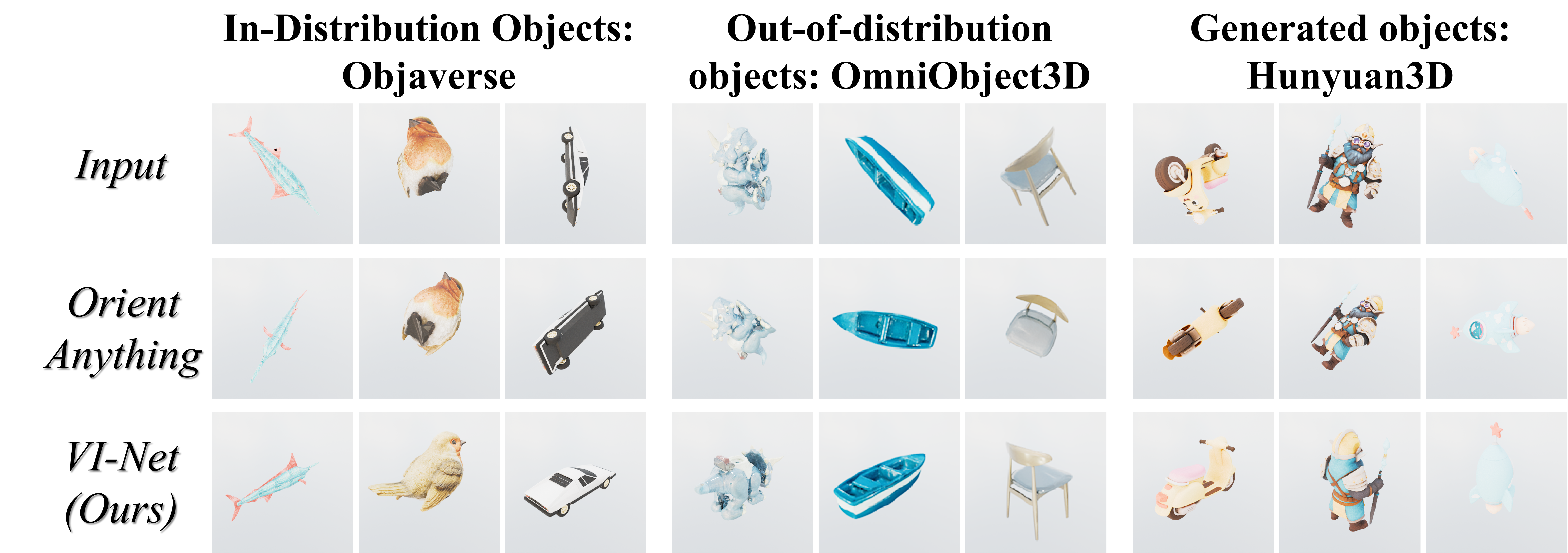}
    \caption{Qualitative comparison of the performance between our model and Orient Anything\cite{orient_anything} on the OmniObject3D\cite{wu2023omniobject3d} dataset shows that the model trained on our dataset has better generalization ability.}
    \label{fig:cod_comapre}
\end{figure}

\paragraph{\textbf{In-Distribution Evaluation.}}
We randomly split 100k samples from our curated dataset for training and 10k samples for held-out testing. 
Based on this split, we retrain multiple state-of-the-art (SOTA) methods, including CleanPose~\cite{lin2025cleanpose}, AGPose~\cite{lin2024instance}, SecondPose~\cite{chen2024secondpose}, and VI-Net~\cite{lin2023vi}, using identical training configurations. 
Detailed hyperparameters are put in the supplementary material.

\begin{table}[t]
    \centering
    \scriptsize
    \caption{In-distribution comparisons for orientation estimation. Best results in \textbf{bold}.}
    \vspace{-3mm}
    \begin{tabular}{c|ccc}
        \toprule
         Method & Acc@$10^\circ \uparrow$ & Acc@$30^\circ \uparrow$ & Abs $\downarrow$\\
         \midrule
         PCA & 22.98 & 34.83 & 91.49 \\
         CleanPose(CanoVerse)\cite{lin2025cleanpose} & 35.02 & 57.36 & 60.72 \\
         AGPose(CanoVerse)\cite{lin2024instance}  & 23.80 & 48.88 & 70.39  \\
         SecondPose(CanoVerse)\cite{chen2024secondpose} & 49.15 & 58.39 & 60.11 \\
         VI-Net(Objaverse-OA)\cite{lin2023vi} & 21.77 & 25.86 & 97.69\\
         VI-Net(CanoVerse)\cite{lin2023vi} & \textbf{55.31} & \textbf{64.14} & \textbf{51.47}  \\
         \bottomrule
    \end{tabular}
    \label{tab:orient_estimation}
    \vspace{-4mm}
\end{table}

\textit{Results.}
As shown in Table~\ref{tab:orient_estimation}, all learning-based methods trained on our dataset outperform the traditional PCA method, demonstrating that geometric cues alone cannot ensure robust canonical alignment for large-scale objects with complex shapes. Among network-based approaches, VI-Net trained on our Canoverse dataset achieves the best performance across all evaluation metrics and excellent generalization over diverse object categories. Compared with models trained on the recent dataset Objaverse-OA\cite{lu2025orientationmatters}, our method shows clear advantages, verifying its effectiveness.

\paragraph{\textbf{Out-of-Distribution Generalization.}}
To evaluate cross-dataset transferability, we test models trained on our dataset on OmniObject3D~\cite{wu2023omniobject3d}, which is reconstructed from real-world scans, serving as an independent out-of-distribution benchmark.
We additionally compare against two mainstream general-purpose 6D pose estimation methods, including GenPose++~\cite{zhang2024omni6dpose} and Orient Anything~\cite{orient_anything}. 
For a fair comparison, we provide the full object point cloud to GenPose++, while for Orient Anything, we use the croplargeEX2 version for experiments, and we render six orthogonal views and select the highest-confidence prediction as the final output.

\textit{Results.}
As summarized in Table~\ref{tab:compare_original}, VI-Net retrained on our dataset significantly outperforms the same model trained on NOCS and Objaverse-OA when evaluated on unseen OOD data.
Furthermore, compared with other large-scale 6D pose estimation methods, it demonstrates superior cross-distribution generalization. These results highlight the advantages of richer category coverage and large-scale canonical annotations.
Figure~\ref{fig:cod_comapre} further confirms that models trained on our dataset produce more consistent and accurate object orientations across in-distribution samples, out-of-distribution objects, and generative outputs. 

\noindent {\emph{\textbf{Scalability Analysis.}}}
To quantify data scaling effects, we train separate models using 32k, 100k, 200k, and 310k training samples, respectively, while keeping the evaluation protocol unchanged.
We train and evaluate using the VI‑Net~\cite{lin2023vi} model and test on the in‑distribution dataset.

\begin{figure}[t]
    \centering
    \includegraphics[width=.8\linewidth]{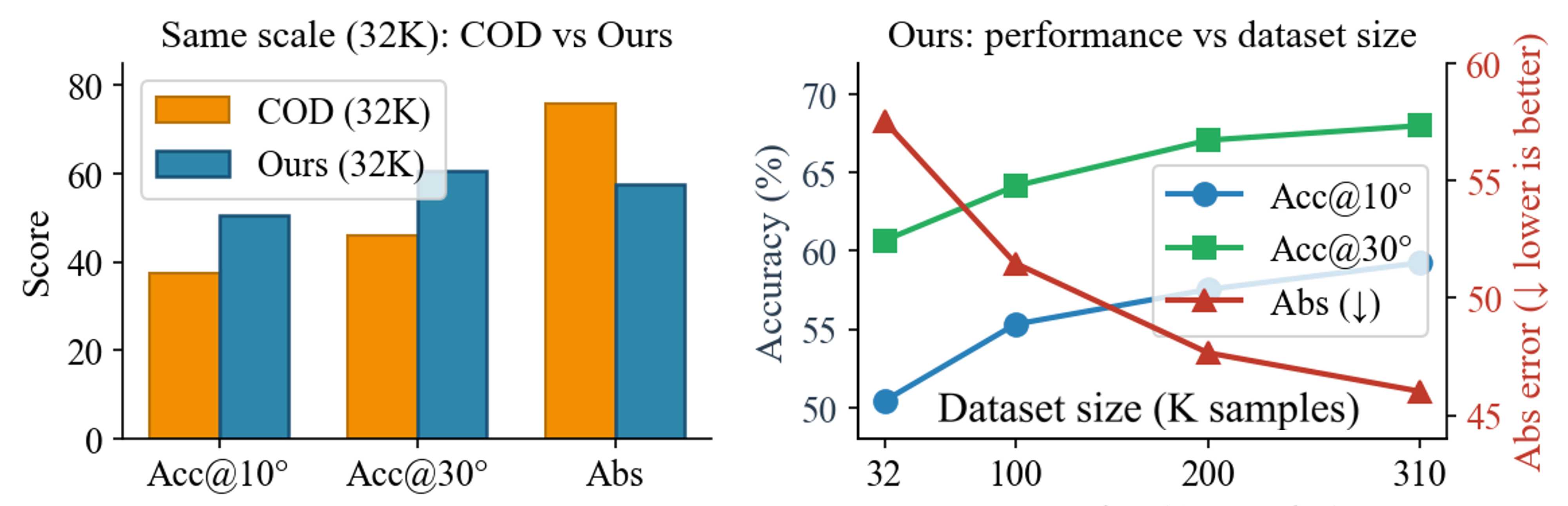}
    \vspace{-3mm}
    \caption{Scalability Analysis on the in-distribution dataset. Left: COD vs. Ours at 32K samples. Right: Ours performance versus dataset size (32K → 310K). For metrics: higher Acc@10° and Acc@30° are better; lower Abs error is better.}
    \label{fig:scaling}
\end{figure}

\textit{Results.}
As shown in Fig.~\ref{fig:scaling}, orientation estimation accuracy improves steadily with increasing training data.
At the same data scale (32K), the model trained on our dataset outperforms its counterpart trained on COD, highlighting the higher quality of our annotations.
This controlled scaling study demonstrates the data efficiency and scalability of our canonical dataset, suggesting that larger-scale 3D datasets can further benefit downstream 3D tasks.

\begin{table}[t]
    \centering
    \scriptsize
    \caption{Out-of-distribution orientation estimation results on the real-world scanned dataset OmniObject3D~\cite{wu2023omniobject3d}.}
    \vspace{-3mm}
    \begin{tabular}{c|ccc}
        \toprule
         Method & Acc@$10^\circ \uparrow$ & Acc@$30^\circ \uparrow$ & Abs $\downarrow$\\
         \midrule
         GenPose++\cite{zhang2024omni6dpose} & 4.33 & 6.95 & 105.17 \\
         Orient Anything\cite{orient_anything} & 4.77 & 7.32 & 103.11 \\
          VI-Net(Nocs)\cite{lin2023vi} & 3.92 & 6.64 & 108.51 \\
          VI-Net(Objaverse-OA)\cite{lin2023vi} & 6.56 & 9.60 & 105.90 \\
        VI-Net(CanoVerse)\cite{lin2023vi} & \textbf{20.18} & \textbf{24.83} & \textbf{94.60} \\
         \bottomrule
    \end{tabular}
    \label{tab:compare_original}
    \vspace{-5mm}
\end{table}


\subsection{3D Object Generation}
\label{sec:3dgen}
To evaluate the impact of canonical data on 3D object generation, we fine-tune two representative generative backbones—Hunyuan3D 2.1~\cite{hunyuan3d2025hunyuan3d} and Trellis~\cite{xiang2024structured}, on 100K samples from CanoVerse, limited by available computational resources.  Specifically, the task follows a straightforward input-output paradigm: input an image and output a generated 3D asset.

\noindent \textbf{Experiment Settings.}
To ensure fair comparison, we conduct strictly controlled single-variable experiments, where data canonicalization is the only varying factor. All fine-tuning runs share identical 3D assets, training configurations, and optimization steps. We directly compare models fine-tuned on raw (non-canonicalized) data against those fine-tuned on canonical CanoVerse data. Evaluation is performed on the public Toys4K~\cite{stojanov21cvpr} and GSO~\cite{downs2022googlescannedobjectshighquality} datasets.


\noindent \textbf{Metrics.}
We evaluate geometric and perceptual consistency using Chamfer Distance (CD)~\cite{zhang2018perceptual}, LPIPS~\cite{zhang2018perceptual}, and CLIP Score~\cite{radford2021learning}. 
Pose stability is measured by the interquartile range (IQR) of angular errors between generated object poses and canonical poses.
CD is computed on 100K mesh points sampled after normalization to [-1,1]. 
For LPIPS and CLIP, meshes are normalized to [-1,1], rendered from six orthogonal views, and averaged across views to evaluate appearance fidelity and orientation alignment. 
For pose stability, we randomly sample 100 objects, obtain their ground-truth canonical poses from our annotation pipeline, compute angular errors with respect to these poses, and report the IQR.

\begin{figure}[t]
    \centering
    \includegraphics[width=.89\linewidth]{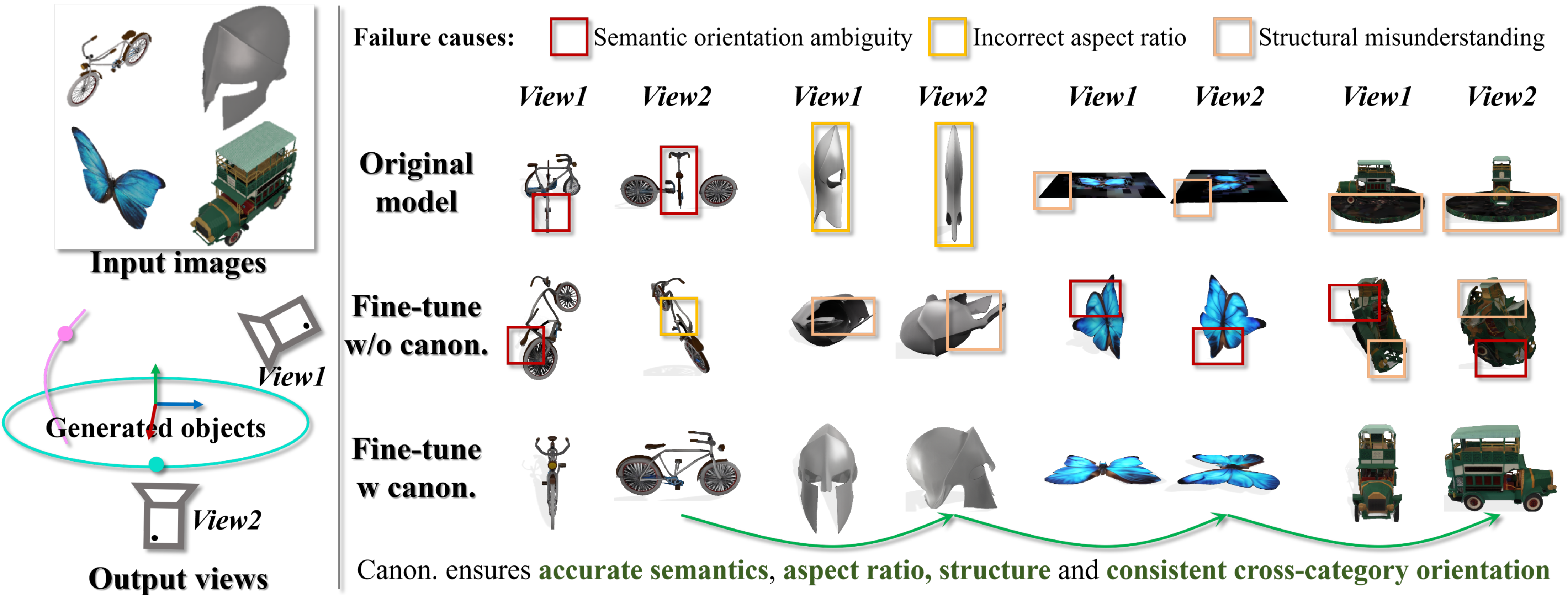}
    \vspace{-2mm}
    \caption{Qualitative analysis of 3D object generation. Fine-tuning with canonical representation outperforms fine-tuning without it in semantics, aspect ratio, and structure.}
    \label{fig:gen_com}
\end{figure}

\noindent \textbf{Results.}
As illustrated in Fig.~\ref{fig:gen_com}, models fine-tuned on canonical data generate shapes with more consistent orientations and structurally coherent geometries. Canonical training reduces pose-induced ambiguity during learning, leading to improved part structure and proportion consistency. Which is also helpful to enhance the model's semantic understanding of object structures, yielding higher-fidelity 3D shapes with eliminated structural ambiguities, artifacts and unreasonable deformations.
Quantitative results in Table~\ref{tab:generate_metric_combined} further confirm that, across both generative backbones, canonical fine-tuning consistently improves orientation alignment, geometric accuracy, and perceptual consistency.
These findings highlight the practical benefits of canonicalized data for stable and high-quality 3D generation.




\begin{table}[t]
    \centering
    \setlength{\tabcolsep}{4pt}  
    \scriptsize
    \vspace{-1mm}
        \caption{
    3D Object Generation comparison results.
    Lower IQR(degree) indicates better pose stability.
    All variants are fine-tuned for the same total training time.
    }
    \vspace{-3mm}
    \begin{tabular}{l c | c c c c| c c c c}  
        \toprule
        & \multicolumn{1}{c|}{} & \multicolumn{4}{c|}{Toys4K } & \multicolumn{4}{c}{GSO } \\  
        \cmidrule(lr){3-6} \cmidrule(lr){7-10}  
        Method & canon. & CD$\downarrow$ & LPIPS$\downarrow$ & CLIP$\uparrow$ & IQR($^\circ$)$\downarrow$ & CD$\downarrow$ & LPIPS$\downarrow$ & CLIP$\uparrow$& IQR($^\circ$)$\downarrow$ \\
        \midrule
        \multirow{2}{*}{Trellis~\cite{xiang2024structured}} 
            & w/o  & 0.227 & 0.436 & 89.7 & 48.5 & 0.272 & 0.468 & 88.0 & 74.0\\
            & w   & \textbf{0.095} & \textbf{0.254} & \textbf{93.9} & \textbf{7.7} & \textbf{0.072} & \textbf{0.244} & \textbf{92.5} &\textbf{8.1}\\
        \midrule
        \multirow{2}{*}{Hun. 2.1~\cite{hunyuan3d2025hunyuan3d}} 
            & w/o  & 0.258 & 0.413 & 90.8 & 75.59 & 0.244 & 0.471 & 88.1 & 74.02\\
            & w   & \textbf{0.085} & \textbf{0.260} & \textbf{93.7}& \textbf{8.1} & \textbf{0.090} & \textbf{0.288} & \textbf{90.6} &\textbf{18.6}\\
        \bottomrule
    \end{tabular}
    \label{tab:generate_metric_combined}
    \vspace{-4mm}
\end{table}



\subsection{Cross-modal 3D Shape Retrieval} \label{sec:shape_ret}
We evaluate two cross-modal 3D shape retrieval tasks: text-to-3D and image-to-3D. 
Given a text or image query, the goal is to retrieve the corresponding 3D instance. 
We examine whether canonicalization reduces directional ambiguity caused by non-canonical training data and improves cross-modal alignment.

\noindent
\textbf{Experimental Setting. }
We randomly sample 310k canonical instances from CanoVerse for training and 10k for testing. 
Each sample forms a triplet of a point cloud, rendered images, and text descriptions generated by Qwen2.5-VL~\cite{qwen2.5-VL}. 
For a strictly controlled comparison, we construct the non-canonical baseline using the same underlying Objaverse assets but apply only scale normalization without canonical alignment, ensuring that canonicalization is the sole varying factor. 
We fine-tune ULIP~\cite{xue2024ulip} and Uni3D~\cite{zhou2023uni3d} under identical configurations. 
In addition to in-distribution evaluation on our test split, we conduct out-of-distribution testing on OmniObject3D~\cite{wu2023omniobject3d}. 
We report Recall@10 and Recall@30 based on exact instance matching. More details are in the supplmentary material.

\begin{table}[htbp]
    \caption{
    Quantitative Comparison of Retrieval Performance.
    Higher Recall@10 and Recall@30 scores indicate better retrieval performance.
    }
    \vspace{-3mm}
    \centering
    \setlength{\tabcolsep}{6pt}  
    \scriptsize
    \begin{tabular}{l c c | c c | c c}  
        \toprule
        & & & \multicolumn{2}{c|}{Canoverse} & \multicolumn{2}{c}{OmniObject3D } \\  
        \cmidrule(lr){4-5} \cmidrule(lr){6-7}  
        Method & Canon. & Modality & Re.@10$\uparrow$ & Re.@30$\uparrow$ & Re.@10$\uparrow$ & Re.@30$\uparrow$ \\
        \midrule
        ULIP\cite{xue2024ulip}       & w/o & Text & 21.33 & 35.17 & 8.19 & 15.98 \\
        Uni3d\cite{zhou2023uni3d}    & w/o & Text & 51.29 & 62.18 & 34.21 & 48.38 \\
        ULIP\cite{xue2024ulip}       & w    & Text & 25.05 & 39.43 & 9.95 & 19.89 \\
        Uni3d\cite{zhou2023uni3d}    & w     & Text & \textbf{53.88} & \textbf{64.31} & \textbf{35.88} & \textbf{49.02} \\
        \noalign{\global\arrayrulewidth1pt}\hline\noalign{\global\arrayrulewidth0.4pt}
        ULIP\cite{xue2024ulip}       & w/o & Image & 19.03 & 32.39 & 8.46 & 16.55 \\
        Uni3d\cite{zhou2023uni3d}    & w/o & Image & 79.07 & 92.29 & 61.34 & 77.42 \\
        ULIP\cite{xue2024ulip}       & w     & Image & 23.63 & 39.04 & 9.66 & 21.50 \\
        Uni3d\cite{zhou2023uni3d}    & w     & Image & \textbf{81.82} & \textbf{92.98} & \textbf{64.86} & \textbf{81.50} \\
        \bottomrule
    \end{tabular}
    \label{tab:all_data_retrieval_combined}
    \vspace{-4mm}
\end{table}




\noindent
\textbf{Results. }
As shown in Table~\ref{tab:all_data_retrieval_combined}, models trained on canonical data consistently outperform those trained on non-canonical data across both ULIP~\cite{xue2024ulip} and Uni3D~\cite{zhou2023uni3d}, for text-to-3D and image-to-3D retrieval on both Canoverse and OmniObject3D~\cite{wu2023omniobject3d}.
These results suggest that canonicalization reduces directional ambiguity and enables models to better capture intrinsic 3D shape semantics, thereby improving cross-modal alignment and generalization.

Canonical annotation methods fall into two categories: (1) \textbf{manual annotation}, which is widely adopted in existing datasets including ShapeNet~\cite{shapenet2015} and OmniObject3D~\cite{wu2023omniobject3d}; (2) \textbf{self-supervised automatic canonicalization methods}, which remain in the research stage (e.g., CaCa~\cite{sun2020canonical}, ConDor~\cite{sajnani2022_condor}) with limited real-world deployment. To validate the effectiveness of our proposed method, we conduct comprehensive comparative experiments against both categories.

\noindent \textbf{Experimental Setup.}
Following COD~\cite{jinli2025}, we conduct all experiments on the DREDS dataset~\cite{dai2022dreds}, which contains seven object categories. All methods are evaluated on objects that are randomly rotated beforehand.

For comparison with \textbf{manual annotation} pipelines, we consider two baselines: Blender-based manual alignment and the COD candidate pose selection pipeline. In COD, four initial random rotations are applied to each object, which are then processed to generate candidate canonical poses, from which annotators select the best one. We evaluate both annotation efficiency and precision. Ten annotators with no prior pose annotation experience are recruited and given standardized training, and their proficiency is evaluated across multiple annotation rounds. See the supplementary material for more details.
For \textbf{automatic annotation} methods, we follow the evaluation protocol of COD~\cite{jinli2025} and adopt two metrics for canonicalization accuracy: Instance-Level Consistency (IC) and Ground Truth Equivariance Consistency (GEC).


\noindent \textbf{Results.}
Compared with manual annotation methods (Table~\ref{tab:annotation_comparison}), 
Blender-based manual canonicalization is intuitive but extremely time-consuming. 
Our method significantly improves efficiency, requiring only 2.6 seconds per object for high-skill annotators compared to 94.5 seconds for manual annotation (\textbf{36× speedup}), and achieving an even larger improvement for low-skill annotators (\textbf{31×}). 
Despite this speedup, our method maintains competitive precision (5.9° for the high-skill group), close to medium-skill manual annotation (4.5°), while significantly outperforming the COD baseline. 
Moreover, training annotators for image selection is substantially easier than learning 3D software operations. 
Considering that the DREDS dataset itself is manually annotated and object alignment can be inherently ambiguous across different geometries, angular errors within 10° are regarded as acceptable systematic fluctuations.

\begin{table}[t]
  \centering
  \scriptsize
  \caption{Annotation Speed and Precision Comparison. Smaller is better for both speed (s/object) and precision (°). Best results are in \textbf{Bold}, and second best is \underline{underlined}.}
  \vspace{-3mm}
  \label{tab:annotation_comparison}
  \setlength{\tabcolsep}{4pt}
  \begin{tabular}{lcccccc}
    \toprule
    \multirow{2}{*}{Method} & \multicolumn{3}{c}{Speed (s/object)} & \multicolumn{3}{c}{Precision (°)} \\
    \cmidrule(lr){2-4} \cmidrule(lr){5-7}
    & Low & Medium & High & Low & Medium & High \\
    \midrule
    Blender (Manual) & 125.3 & 101.4 & 94.5 & \underline{8.2} & \textbf{4.5} & \textbf{3.7} \\
    COD~\cite{jinli2025} & \underline{4.6} & \underline{3.4} & \underline{3.0} & 9.4 & 8.5 & 8.1 \\
    Ours & \textbf{4.0} & \textbf{2.7} & \textbf{2.6} & \textbf{6.7} & \underline{6.3} & \underline{5.9} \\
    \bottomrule
  \end{tabular}
\end{table}

\subsection{Annotation Quality and Efficiency} \label{sec:anno_val}

\begin{table*}[t]
\centering
\caption{Comparison of 3D object canonicalization on the DREDS~\cite{dai2022dreds} dataset.}
\vspace{-3mm}
\label{tab:canonicalization_compare}
\resizebox{\textwidth}{!}{ 
\begin{tabular}{l|cccccccccccccc|cc}
\toprule
\multirow{2}{*}{Method}
& \multicolumn{2}{c}{Aeroplane} 
& \multicolumn{2}{c}{Car} 
& \multicolumn{2}{c}{Bowl} 
& \multicolumn{2}{c}{Camera}  
& \multicolumn{2}{c}{Bottle}  
& \multicolumn{2}{c}{Can} 
& \multicolumn{2}{c}{Mug} 
& \multicolumn{2}{c}{Avg} \\

\cmidrule(lr){2-3} \cmidrule(lr){4-5} \cmidrule(lr){6-7} \cmidrule(lr){8-9}  
\cmidrule(lr){10-11} \cmidrule(lr){12-13} \cmidrule(lr){14-15} \cmidrule(lr){16-17} 
& IC & GEC & IC & GEC & IC & GEC & IC & GEC  
& IC & GEC & IC & GEC & IC & GEC & IC & GEC \\  
\midrule

CaCa\cite{sun2020canonical} & 0.785 & 0.869 & 1.985 & 1.881 & 1.569 & 1.895 & 1.294 & 1.451 & 2.632 & 3.080 & 1.797 & 2.095 & 0.661 & 0.655 & 1.532 & 1.704 \\  
ConDor\cite{sajnani2022_condor} & 0.823 & 1.391 & 0.605 & 0.782 & 0.898 & 1.202 & 1.035 & 1.529 & 1.108 & 1.261 & 1.048 & 1.753 & 0.126 & 0.189 & 0.806 & 1.158 \\  
ShapeMat.\cite{di2024shapematcher} & 0.659 & 2.337 & 2.409 & 3.596 & 0.368 & 0.452 & 0.377 & 1.491 & 1.279 & 3.791 & 1.369 & 2.211 & 0.399 & 0.629 & 0.980 & 2.072 \\  
COD\cite{jinli2025} &  \textbf{0.051} & \textbf{0.058} & 0.103 & 0.118 & \textbf{0.011} & \textbf{0.012} & 1.116 & 1.177 & 0.031 & 0.034 & 0.030 & 0.037 & \textbf{0.018} & \textbf{0.019} & 0.194 & 0.208 \\  
Ours &0.129 & 0.176 & \textbf{0.007} & \textbf{0.012} & 0.018 & 0.032 & \textbf{0.033} & \textbf{0.584} & \textbf{0.009} & \textbf{0.012} & \textbf{0.005} &  \textbf{0.017} & 0.022 & 0.026 & \textbf{0.032} & \textbf{0.123} \\
\bottomrule
\end{tabular}
}

\end{table*}

Against automatic annotation methods (Table~\ref{tab:canonicalization_compare}), 
our method achieves strong overall performance, improving IC by 83.5\% and GEC by 40.9\% over the state-of-the-art method, benefiting from the more reliable candidate poses generated by our pipeline. 
For a small number of categories, our precision is slightly lower than COD due to the difficulty of capturing subtle angular differences when projecting 3D objects to 2D renderings. 
Nevertheless, the errors remain within the acceptable fluctuation range, and the resulting annotations are valid canonical data for downstream tasks.

\section{Conclusions}
We have addressed a persistent bottleneck in 3D vision: the absence of large-scale canonical orientation. We introduced \methodName{}, a 320K/1,156-category canonical 3D dataset, and demonstrated that canonical orientation -- when provided at dataset scale -- emerges as a useful structural prior for modern 3D models. 
\methodName{} not only improves pose stability in 3D generation , but also enables zero-shot orientation estimation from point clouds, a capability that was largely inaccessible before.
This dataset was made possibly by a new canonicalization engine that treats alignment as scalable hypothesis triage instead of exhaustive SO(3) search, reducing human oversight to seconds per object while maintaining high alignment fidelity.
We hope \methodName{} will serve as a new foundation for orientation-consistent 3D learning, benchmarking, and generative modeling.

%
%
\bibliographystyle{splncs04}
\bibliography{main}

@String(CVPR  = {IEEE Conf. Comput. Vis. Pattern Recog.})

@String(ICCV  = {Int. Conf. Comput. Vis.})

@String(ECCV  = {Eur. Conf. Comput. Vis.})

@String(ICLR  = {Int. Conf. Learn. Represent.})

@String(TOG   = {ACM Trans. Graph.})

@String(CVPR  = {CVPR})

@String(ICCV  = {ICCV})

@String(ECCV  = {ECCV})

@String(ICLR  = {ICLR})

@String(TOG   = {ACM TOG})

@misc{lu2025orientationmatters,
      title={Orientation Matters: Making 3D Generative Models Orientation-Aligned}, 
      author={Yichong Lu and Yuzhuo Tian and Zijin Jiang and Yikun Zhao and Yuanbo Yang and Hao Ouyang and Haoji Hu and Huimin Yu and Yujun Shen and Yiyi Liao},
      year={2025},
      eprint={2506.08640},
      archivePrefix={arXiv},
      primaryClass={cs.CV},
      url={https://arxiv.org/abs/2506.08640}, 
}

@InProceedings{jinli2025,
    author    = {Jin, Li and Wang, Yujie and Chen, Wenzheng and Dai, Qiyu and Gao, Qingzhe and Qin, Xueying and Chen, Baoquan},
    title     = {One-shot 3D Object Canonicalization based on Geometric and Semantic Consistency},
    booktitle = {CVPR},
    month     = {June},
    year      = {2025},
    pages     = {16850-16859}
}

@inproceedings{wu2023omniobject3d,
    author = {Tong Wu and Jiarui Zhang and Xiao Fu and Yuxin Wang and Jiawei Ren, 
    Liang Pan and Wayne Wu and Lei Yang and Jiaqi Wang and Chen Qian and Dahua Lin and Ziwei Liu},
    title = {OmniObject3D: Large-Vocabulary 3D Object Dataset for Realistic Perception, 
    Reconstruction and Generation},
    booktitle={IEEE/CVF Conference on Computer Vision and Pattern Recognition (CVPR)},
    year={2023}
}

@techreport{shapenet2015,
  title       = {{ShapeNet: An Information-Rich 3D Model Repository}},
  author      = {Chang, Angel X. and Funkhouser, Thomas and Guibas, Leonidas and Hanrahan, Pat and Huang, Qixing and Li, Zimo and Savarese, Silvio and Savva, Manolis and Song, Shuran and Su, Hao and Xiao, Jianxiong and Yi, Li and Yu, Fisher},
  number      = {arXiv:1512.03012 [cs.GR]},
  institution = {Stanford University --- Princeton University --- Toyota Technological Institute at Chicago},
  year        = {2015}
}

@inproceedings{Zhirong15CVPR,
    Author = {Wu, Zhirong and Song, Shuran and Khosla, Aditya and Yu, Fisher and Zhang, Linguang and Tang, Xiaoou and Xiao, Jianxiong},
    Title = {3D ShapeNets: A Deep Representation for Volumetric Shapes},
    Booktitle = {{IEEE} Conference on Computer Vision and Pattern Recognition (CVPR)},
    Year = {2015}
}

@inproceedings{wen2024foundationpose,
  title={Foundationpose: Unified 6d pose estimation and tracking of novel objects},
  author={Wen, Bowen and Yang, Wei and Kautz, Jan and Birchfield, Stan},
  booktitle={Proceedings of the IEEE/CVF Conference on Computer Vision and Pattern Recognition},
  pages={17868--17879},
  year={2024}
}

@inproceedings{peng2019pvnet,
  title={Pvnet: Pixel-wise voting network for 6dof pose estimation},
  author={Peng, Sida and Liu, Yuan and Huang, Qixing and Zhou, Xiaowei and Bao, Hujun},
  booktitle={Proceedings of the IEEE/CVF conference on computer vision and pattern recognition},
  pages={4561--4570},
  year={2019}
}

@inproceedings{sun2022onepose,
  title={Onepose: One-shot object pose estimation without cad models},
  author={Sun, Jiaming and Wang, Zihao and Zhang, Siyu and He, Xingyi and Zhao, Hongcheng and Zhang, Guofeng and Zhou, Xiaowei},
  booktitle={Proceedings of the IEEE/CVF Conference on Computer Vision and Pattern Recognition},
  pages={6825--6834},
  year={2022}
}

@inproceedings{lee2025any6d,
  title={Any6D: Model-free 6D Pose Estimation of Novel Objects},
  author={Lee, Taeyeop and Wen, Bowen and Kang, Minjun and Kang, Gyuree and Kweon, In So and Yoon, Kuk-Jin},
  booktitle={Proceedings of the Computer Vision and Pattern Recognition Conference},
  pages={11633--11643},
  year={2025}
}

@inproceedings{wang2019normalized,
  title={Normalized object coordinate space for category-level 6d object pose and size estimation},
  author={Wang, He and Sridhar, Srinath and Huang, Jingwei and Valentin, Julien and Song, Shuran and Guibas, Leonidas J},
  booktitle={Proceedings of the IEEE/CVF conference on computer vision and pattern recognition},
  pages={2642--2651},
  year={2019}
}

@inproceedings{chen2020category,
  title={Category level object pose estimation via neural analysis-by-synthesis},
  author={Chen, Xu and Dong, Zijian and Song, Jie and Geiger, Andreas and Hilliges, Otmar},
  booktitle={European Conference on Computer Vision},
  pages={139--156},
  year={2020},
  organization={Springer}
}

@inproceedings{lin2022category,
  title={Category-level 6d object pose and size estimation using self-supervised deep prior deformation networks},
  author={Lin, Jiehong and Wei, Zewei and Ding, Changxing and Jia, Kui},
  booktitle={European Conference on Computer Vision},
  pages={19--34},
  year={2022},
  organization={Springer}
}

@inproceedings{fan2022object,
  title={Object level depth reconstruction for category level 6d object pose estimation from monocular rgb image},
  author={Fan, Zhaoxin and Song, Zhenbo and Xu, Jian and Wang, Zhicheng and Wu, Kejian and Liu, Hongyan and He, Jun},
  booktitle={European Conference on Computer Vision},
  pages={220--236},
  year={2022},
  organization={Springer}
}

@inproceedings{zhou2023uni3d,
  title={Uni3d: Exploring unified 3d representation at scale},
  author={Zhou, Junsheng and Wang, Jinsheng and Ma, Baorui and Liu, Yu-Shen and Huang, Tiejun and Wang, Xinlong},
  booktitle={International Conference on Learning Representations (ICLR)},
  year={2024}
}

@misc{ma20243d,
      title={Find Any Part in 3D}, 
      author={Ziqi Ma and Yisong Yue and Georgia Gkioxari},
      year={2024},
      eprint={2411.13550},
      archivePrefix={arXiv},
      primaryClass={cs.CV},
      url={https://arxiv.org/abs/2411.13550}, 
}

@misc{hunyuan3d2025hunyuan3d,
    title={Hunyuan3D 2.1: From Images to High-Fidelity 3D Assets with Production-Ready PBR Material},
    author={Tencent Hunyuan3D Team},
    year={2025},
    eprint={2506.15442},
    archivePrefix={arXiv},
    primaryClass={cs.CV}
}

@inproceedings{lin2023vi,
  title={Vi-net: Boosting category-level 6d object pose estimation via learning decoupled rotations on the spherical representations},
  author={Lin, Jiehong and Wei, Zewei and Zhang, Yabin and Jia, Kui},
  booktitle={Proceedings of the IEEE/CVF International Conference on Computer Vision},
  pages={14001--14011},
  year={2023}
}

@inproceedings{lin2025cleanpose,
  title={Cleanpose: Category-level object pose estimation via causal learning and knowledge distillation},
  author={Lin, Xiao and Peng, Yun and Wang, Liuyi and Zhong, Xianyou and Zhu, Minghao and Yang, Jingwei and Feng, Yi and Liu, Chengju and Chen, Qijun},
  booktitle={Proceedings of the IEEE/CVF International Conference on Computer Vision (ICCV)},
  year={2025}
}

@inproceedings{chen2024secondpose,
  title={Secondpose: Se (3)-consistent dual-stream feature fusion for category-level pose estimation},
  author={Chen, Yamei and Di, Yan and Zhai, Guangyao and Manhardt, Fabian and Zhang, Chenyangguang and Zhang, Ruida and Tombari, Federico and Navab, Nassir and Busam, Benjamin},
  booktitle={Proceedings of the IEEE/CVF Conference on Computer Vision and Pattern Recognition},
  pages={9959--9969},
  year={2024}
}

@inproceedings{lin2024instance,
  title={Instance-adaptive and geometric-aware keypoint learning for category-level 6d object pose estimation},
  author={Lin, Xiao and Yang, Wenfei and Gao, Yuan and Zhang, Tianzhu},
  booktitle={Proceedings of the IEEE/CVF Conference on Computer Vision and Pattern Recognition},
  pages={21040--21049},
  year={2024}
}

@article{stojanov21cvpr,
      title={Using Shape to Categorize: Low-Shot Learning with an Explicit Shape Bias},
      author={Stefan Stojanov and Anh Thai and James M. Rehg},
      booktitle = {CVPR},
      year      = {2021}
}

@inproceedings{zhang2024omni6dpose,
  title={Omni6dpose: A benchmark and model for universal 6d object pose estimation and tracking},
  author={Zhang, Jiyao and Huang, Weiyao and Peng, Bo and Wu, Mingdong and Hu, Fei and Chen, Zijian and Zhao, Bo and Dong, Hao},
  booktitle={European Conference on Computer Vision},
  pages={199--216},
  year={2024},
  organization={Springer}
}

@misc{qwen2.5-VL,
    title = {Qwen2.5-VL},
    url = {https://qwenlm.github.io/blog/qwen2.5-vl/},
    author = {Qwen Team},
    month = {January},
    year = {2025}
}

@inproceedings{xue2024ulip,
  title={Ulip-2: Towards scalable multimodal pre-training for 3d understanding},
  author={Xue, Le and Yu, Ning and Zhang, Shu and Panagopoulou, Artemis and Li, Junnan and Mart{\'\i}n-Mart{\'\i}n, Roberto and Wu, Jiajun and Xiong, Caiming and Xu, Ran and Niebles, Juan Carlos and others},
  booktitle={Proceedings of the IEEE/CVF Conference on Computer Vision and Pattern Recognition},
  pages={27091--27101},
  year={2024}}

@article{Upright08,
  author       = {Hongbo Fu and
                  Daniel Cohen{-}Or and
                  Gideon Dror and
                  Alla Sheffer},
  title        = {Upright orientation of man-made objects},
  journal      = {{ACM} Trans. Graph.},
  volume       = {27},
  number       = {3},
  pages        = {42},
  year         = {2008},
  url          = {https://doi.org/10.1145/1360612.1360641},
  doi          = {10.1145/1360612.1360641},
  bibsource    = {dblp computer science bibliography, https://dblp.org}
}

@article{wu2025direct3d,
  title={Direct3d-s2: Gigascale 3d generation made easy with spatial sparse attention},
  author={Wu, Shuang and Lin, Youtian and Zhang, Feihu and Zeng, Yifei and Yang, Yikang and Bao, Yajie and Qian, Jiachen and Zhu, Siyu and Cao, Xun and Torr, Philip and others},
  journal={arXiv preprint arXiv:2505.17412},
  year={2025}
}

@article{trellis,
  title={Structured 3d latents for scalable and versatile 3d generation},
  author={Xiang, Jianfeng and Lv, Zelong and Xu, Sicheng and Deng, Yu and Wang, Ruicheng and Zhang, Bowen and Chen, Dong and Tong, Xin and Yang, Jiaolong},
  journal={arXiv preprint arXiv:2412.01506},
  year={2024}
}

@article{hunyuan2,
  title={Hunyuan3d 2.0: Scaling diffusion models for high resolution textured 3d assets generation},
  author={Zhao, Zibo and Lai, Zeqiang and Lin, Qingxiang and Zhao, Yunfei and Liu, Haolin and Yang, Shuhui and Feng, Yifei and Yang, Mingxin and Zhang, Sheng and Yang, Xianghui and others},
  journal={arXiv preprint arXiv:2501.12202},
  year={2025}
}

@inproceedings{rombach2022high,
  title={High-resolution image synthesis with latent diffusion models},
  author={Rombach, Robin and Blattmann, Andreas and Lorenz, Dominik and Esser, Patrick and Ommer, Bj{\"o}rn},
  booktitle={Proceedings of the IEEE/CVF conference on computer vision and pattern recognition},
  pages={10684--10695},
  year={2022}
}

@article{dora,
  title={Dora: Sampling and Benchmarking for 3D Shape Variational Auto-Encoders},
  author={Chen, Rui and Zhang, Jianfeng and Liang, Yixun and Luo, Guan and Li, Weiyu and Liu, Jiarui and Li, Xiu and Long, Xiaoxiao and Feng, Jiashi and Tan, Ping},
  journal={arXiv preprint arXiv:2412.17808},
  year={2024}
}

@article{clay,
  title={CLAY: A Controllable Large-scale Generative Model for Creating High-quality 3D Assets},
  author={Zhang, Longwen and Wang, Ziyu and Zhang, Qixuan and Qiu, Qiwei and Pang, Anqi and Jiang, Haoran and Yang, Wei and Xu, Lan and Yu, Jingyi},
  journal={ACM Transactions on Graphics (TOG)},
  volume={43},
  number={4},
  pages={1--20},
  year={2024},
  publisher={ACM New York, NY, USA}
}

@inproceedings{xcubes,
  title={Xcube: Large-scale 3d generative modeling using sparse voxel hierarchies},
  author={Ren, Xuanchi and Huang, Jiahui and Zeng, Xiaohui and Museth, Ken and Fidler, Sanja and Williams, Francis},
  booktitle={Proceedings of the IEEE/CVF conference on computer vision and pattern recognition},
  pages={4209--4219},
  year={2024}
}

@article{craftsman,
  title={Craftsman: High-fidelity mesh generation with 3d native generation and interactive geometry refiner},
  author={Li, Weiyu and Liu, Jiarui and Chen, Rui and Liang, Yixun and Chen, Xuelin and Tan, Ping and Long, Xiaoxiao},
  journal={arXiv preprint arXiv:2405.14979},
  year={2024}
}

@article{3dshape2vecset,
  title={3dshape2vecset: A 3d shape representation for neural fields and generative diffusion models},
  author={Zhang, Biao and Tang, Jiapeng and Niessner, Matthias and Wonka, Peter},
  journal={ACM Transactions On Graphics (TOG)},
  volume={42},
  number={4},
  pages={1--16},
  year={2023},
  publisher={ACM New York, NY, USA}
}

@article{wu2024direct3d,
  title={Direct3d: Scalable image-to-3d generation via 3d latent diffusion transformer},
  author={Wu, Shuang and Lin, Youtian and Zhang, Feihu and Zeng, Yifei and Xu, Jingxi and Torr, Philip and Cao, Xun and Yao, Yao},
  journal={Advances in Neural Information Processing Systems},
  volume={37},
  pages={121859--121881},
  year={2024}
}

@article{xiang2024structured,
    title   = {Structured 3D Latents for Scalable and Versatile 3D Generation},
    author  = {Xiang, Jianfeng and Lv, Zelong and Xu, Sicheng and Deng, Yu and Wang, Ruicheng and Zhang, Bowen and Chen, Dong and Tong, Xin and Yang, Jiaolong},
    journal = {arXiv preprint arXiv:2412.01506},
    year    = {2024}
}

@inproceedings{zhang2018perceptual,
  title={The Unreasonable Effectiveness of Deep Features as a Perceptual Metric},
  author={Zhang, Richard and Isola, Phillip and Efros, Alexei A and Shechtman, Eli and Wang, Oliver},
  booktitle={CVPR},
  year={2018}
}

@inproceedings{radford2021learning,
  title={Learning transferable visual models from natural language supervision},
  author={Radford, Alec and Kim, Jong Wook and Hallacy, Chris and Ramesh, Aditya and Goh, Gabriel and Agarwal, Sandhini and Sastry, Girish and Askell, Amanda and Mishkin, Pamela and Clark, Jack and others},
  booktitle={International conference on machine learning},
  pages={8748--8763},
  year={2021},
  organization={PmLR}
}

@article{orient_anything,
  title={Orient Anything: Learning Robust Object Orientation Estimation from Rendering 3D Models},
  author={Wang, Zehan and Zhang, Ziang and Pang, Tianyu and Du, Chao and Zhao, Hengshuang and Zhao, Zhou},
  journal={arXiv:2412.18605},
  year={2024}
}

@inproceedings{dai2022dreds,
	title={Domain Randomization-Enhanced Depth Simulation and Restoration for Perceiving and Grasping Specular and Transparent Objects},
	author={Dai, Qiyu and Zhang, Jiyao and Li, Qiwei and Wu, Tianhao and Dong, Hao and Liu, Ziyuan and Tan, Ping and Wang, He},
	booktitle={European Conference on Computer Vision (ECCV)},
	year={2022}
    }

@InProceedings{sajnani2022_condor,
author = {Rahul Sajnani and
               Adrien Poulenard and
               Jivitesh Jain and
               Radhika Dua and
               Leonidas J. Guibas and
               Srinath Sridhar},
title = {ConDor: Self-Supervised Canonicalization of 3D Pose for Partial Shapes},
booktitle = {The IEEE Conference on Computer Vision and Pattern Recognition (CVPR)},
month = {June},
year = {2022}
}

@conference{sun2020canonical,
   title={Canonical Capsules: Self-Supervised Capsules in Canonical Pose},
   author={Weiwei Sun and Andrea Tagliasacchi and Boyang Deng and 
           Sara Sabour and Soroosh Yazdani and Geoffrey Hinton and
           Kwang Moo Yi},
   booktitle={Neural Information Processing Systems},
   year={2021}
}

@inproceedings{di2024shapematcher,
  title={Shapematcher: Self-supervised joint shape canonicalization segmentation retrieval and deformation},
  author={Di, Yan and Zhang, Chenyangguang and Wang, Chaowei and Zhang, Ruida and Zhai, Guangyao and Li, Yanyan and Fu, Bowen and Ji, Xiangyang and Gao, Shan},
  booktitle={Proceedings of the IEEE/CVF Conference on Computer Vision and Pattern Recognition},
  pages={21017--21028},
  year={2024}
}

@misc{downs2022googlescannedobjectshighquality,
      title={Google Scanned Objects: A High-Quality Dataset of 3D Scanned Household Items}, 
      author={Laura Downs and Anthony Francis and Nate Koenig and Brandon Kinman and Ryan Hickman and Krista Reymann and Thomas B. McHugh and Vincent Vanhoucke},
      year={2022},
      eprint={2204.11918},
      archivePrefix={arXiv},
      primaryClass={cs.RO},
      url={https://arxiv.org/abs/2204.11918}, 
}
\end{document}